\documentclass{article} 
\usepackage{iclr2027_conference,times}

\usepackage{amsmath,amsfonts,bm}

\def\eqref#1{equation~\ref{#1}}

\def\1{\bm{1}}

\DeclareMathAlphabet{\mathsfit}{\encodingdefault}{\sfdefault}{m}{sl}
\SetMathAlphabet{\mathsfit}{bold}{\encodingdefault}{\sfdefault}{bx}{n}

\usepackage[utf8]{inputenc} 
\usepackage[T1]{fontenc}    
\usepackage{hyperref}       
\usepackage{url}            
\usepackage{booktabs}       
\usepackage{amsfonts}       
\usepackage{nicefrac}       
\usepackage{microtype}      
\usepackage{xcolor}         
\usepackage{amsmath}
\usepackage{graphicx}

\usepackage{tcolorbox}
\usepackage{xcolor}
\usepackage{tabularx}
\tcbuselibrary{skins}
\usepackage{subcaption}
\usepackage{tocbibind}  
\usepackage{titletoc}   
\usepackage{enumitem}
\usepackage{caption}
\usepackage {wrapfig}
\usepackage{minted}
\usepackage{amsthm}
\usepackage{bm}

\tcbuselibrary{breakable}
\usepackage{multirow}
\usepackage{multicol}
\usepackage{colortbl}
\definecolor{my_green}{RGB}{51,102,0}
\definecolor{my_purple}{RGB}{160, 43, 147}
\definecolor{my_blue}{RGB}{15, 158, 213}
\usepackage{fontawesome5}
\definecolor{high}{HTML}{6AD4DD}
\definecolor{math}{HTML}{FFA09B}
\definecolor{darkpink}{RGB}{255, 20, 147}
\newtcolorbox{promptbox}[1]{enhanced,breakable,colback=gray!3,colframe=gray!45,boxrule=0.5pt,arc=1mm,title={#1},colbacktitle=gray!12,coltitle=black,fonttitle=\small\bfseries,fontupper=\small\ttfamily,before upper={\raggedright},left=2mm,right=2mm,top=2mm,bottom=2mm}
\newcommand{\promptitem}[2]{\par\smallskip\noindent\textbf{#1.} #2\par}
\newcommand{\SMART}{\texttt{SMaRT}}
\iclrfinalcopy

\title{Imprint Reader: From Weight-Update Readout to Behavioral Intervention}

\author{
\textbf{Guanxu Chen}$^{1,2}\thanks{Equal Contribution}$\quad
\textbf{Qihao Lin}$^{1,2}$$^{\ast}$\quad
\textbf{Jing Shao}$^{2}\thanks{Corresponding Author.}$\quad
\vspace{1em}\\
$^1$ Shanghai Artificial Intelligence Laboratory,
$^2$ Shanghai Jiao Tong University, \\
\it\footnotesize ~~lm.cgx@sjtu.edu.cn\quad\quad shaojing@pjlab.org.cn
\vspace{1em}\\
~~Our code: \textcolor{darkpink}{\faGithub \href{https://github.com/biuboomc/CANON}{SMaRT}} ~~ Our model: \textcolor{orange}{\faBook \href{https://huggingface.co/quantumfr/imprint-reader-v1.0-0928}{Imprint Reader}}
}

\begin{document}

\maketitle

\begin{abstract}
As language models take a growing role in AI development, a
natural aspiration is for them to reflect on their own learning
process, as humans do, and use that reflection to improve
themselves. At the same time, these models have an advantage that human
learners lack, since training leaves parameter-level traces that
can, in principle, be inspected directly. However, current models 
cannot decode these traces into an explicit account of what they
have learned. To this end, we introduce the \textit{Imprint Reader}, a model trained with \textit{Semantic Mount-and-Read Tuning} (\SMART{}) to describe frozen weight updates. \SMART{} mounts each update onto the Reader and uses an anchor-free meta-query to elicit a natural-language description, while no-change and random-perturbation controls discourage unsupported claims. On held-out updates, the joint Reader reaches judge-based Pass@100 of $2\%$ for knowledge and $16\%$ for behavior. These results demonstrate the feasibility of natural-language readout while pointing to reliability across updates as the next step. Beyond free-form generation, the Reader provides a differentiable proxy for the gap between a specified target behavior and a candidate weight update. Its coordinate-aligned gradients support intervention through MetaEdit. At a $0.5\%$ pruning rate, Reader-guided selection raises
measured harmful-prompt refusal from $57.9\%$ to $64.1\%$
under a safety-maintenance target. Using behavior descriptions without target-task training data, MetaEdit increases the frequency of backtracking and sub-goal expressions in mathematical reasoning traces and raises BFCL Overall from $41.69\%$ to $44.60\%$.
\end{abstract}
\section{Introduction}
\label{sec:intro}

With steadily improving engineering and research capabilities,
language models are taking an ever larger part in the development of
AI itself, from generating training data to writing and reviewing
research code~\citep{wang2023selfinstruct,lu2026aiscientist, zhang2025darwin}. A natural aspiration behind this trend is to remove the human from
the loop entirely, allowing models to reflect on what they have
learned, as human students do, and use that reflection to close
the cycle of self-improvement \citep{good1965speculations,schmidhuber2007godel}. In this task, models have an
advantage over human students because their learning is fully
materialized in their parameters, and every update is open to direct inspection.

However, this
advantage has so far gone unexploited, as today's models can neither
perceive their own learning the way humans do nor read the updates
they physically possess. What is missing is the capacity to decode a weight update into an
explicit account of what the model learned or how its behavior
changed. Recent studies have begun to probe this capacity, but
their readouts rarely provide a specific and reliable account of
what an update changed. Weight-space methods
predict only coarse attributes such as accuracy or the fine-tuning
task~\citep{unterthiner2020predicting, schurholt2021self,
eilertsen2020classifying, putterman2025learning, han2026w2t}, while
the few that verbalize weight differences are confined to narrow,
purpose-built domains and readily fabricate descriptions for updates
that carry no information~\citep{goel2026learning,
shenoy2026introspection}. In either case, the readout ends at
monitoring and offers no path toward acting on what is decoded.

To this end, we invert the usual direction of weight readout. Rather
than attaching an adaptor to each fine-tuned model and asking it to
describe itself, we train a single complete model, the
\textit{Imprint Reader}, which mounts a frozen weight update onto its
own parameters and describes the factual knowledge or behavioral
change associated with that update. Because the Reader shares
parameter coordinates with its parent, its gradients live in the
same space as the parent's parameters, turning readout from passive
monitoring into a natural interface for intervention. Specifically,
we construct weight updates from examples designed to induce either
factual knowledge or a behavioral tendency, and optimize the Reader
with \textit{Semantic Mount-and-Read Tuning} (\SMART{}) to describe the change
associated with each update in natural language. The Reader is
prompted only by an anchor-free meta-query sampled independently of
the target change, so the query provides no item-specific cue about
what was learned. We further design paired control episodes with
empty or random updates, training the Reader to abstain rather than
fabricate when an update carries no recoverable semantics.

Empirically, we establish both the feasibility and current limits
of reading newly acquired knowledge and behavior from weight
updates. We train a single Reader jointly on knowledge-bearing
and behavior-inducing updates from the Qwen3-14B \citep{yang2025qwen3}.
The Reader's generated
descriptions reach judge-based Pass@100 of $2\%$ for knowledge
and $16\%$ for behavior under anchor-free meta-queries. These
results show that natural-language readout is feasible, while
its reliability across updates remains to be improved.

Beyond free-form readout, the Reader provides a differentiable
proxy for the gap between a specified target behavior and a candidate weight update. Its coordinate-aligned
gradients thus enable MetaEdit to intervene on the original
Qwen3-14B. At a $0.5\%$ pruning rate, Reader-guided row pruning
shifts the measured harmful-prompt refusal rate from $57.9\%$ to
$64.1\%$ under a safety-maintenance target and to $55.4\%$
under a refusal-relaxation target. In mathematics, sparse updates increase the frequency of
backtracking and sub-goal expressions in generated reasoning
traces. On BFCL, they raise Overall from $41.69\%$ to $44.60\%$,
using behavior descriptions but no training examples from either
target task. We call this description-driven intervention
\emph{vibe alignment}.

Overall, these results suggest that weight updates can provide
signals for both natural-language readout and targeted intervention. The Reader can
describe factual and behavioral changes from updates it has not
seen, and its coordinate-aligned gradients allow MetaEdit to
act on the original model using descriptions of desired behavior
without target-task training data. Although the reliability
of natural-language readout remains to be improved, the
intervention results show that a complete generated description
is not required to use the Reader's parameter-space signal. We
view this readout-and-intervention interface as an initial step
toward models that can inspect, verify, and eventually adjust
their own learning process.
\section{Related Work}
\textbf{Reading Neural Network Weights.} Weight-space learning treats
model parameters as a data modality and trains external predictors
over them~\citep{wang2026wsl}. Early studies show that model weights retain information about
training and performance. \citet{unterthiner2020predicting} predict
test accuracy from weights, \citet{eilertsen2020classifying} infer
training hyperparameters such as the optimizer and batch size, and
\citet{schurholt2021self} learn self-supervised weight embeddings
that transfer to model-property prediction. A parallel
line designs architectures that respect parameter symmetries,
including permutation-equivariant networks for MLP and CNN
weights~\citep{navon2023equivariant, zhou2023permutation}, their
extensions to general architectures~\citep{zhou2024universal}, and
graph-based metanetworks that process heterogeneous
models~\citep{lim2024graph, kofinas2024graph}. Beyond property prediction, \citet{haim2022reconstructing}
reconstruct training samples from model weights. Recent work also
classifies fine-tuning tasks from LoRA weights~\citep{putterman2025learning}
and predicts the capabilities conferred by an adapter~\citep{han2026w2t}.
Our focus is the natural-language description of specific factual
and behavioral changes from held-out weight updates.

\textbf{Model Introspection and Self-Description.}
Another line studies whether language models can report on their
own knowledge, behavior, and internal states.
\citet{kadavath2022language} study whether models can assess when
they know an answer, while \citet{lin2022teaching} train models to
express uncertainty in words. \citet{binder2024looking} examine
models' predictions of their own behavior, and
\citet{laine2024sad} benchmark situational self-knowledge.
Intermediate representations have also been decoded into natural
language~\citep{chen2024selfie,ghandeharioun2024patchscopes}, while
concept injection has been used to probe models' access to their
own activations~\citep{lindsey2025emergent}.

Closer to weight-update readout, \citet{betley2025tell} find partial
awareness of learned behaviors in fine-tuned models,
\citet{goel2026learning} train adapters to describe the behavioral
effects of weight differences, and \citet{shenoy2026introspection}
study such readout across multiple models. Our setting additionally
tests recovery of specific factual propositions under anchor-free
meta-queries, includes no-change and random-perturbation controls
for abstention, and uses coordinate-aligned Reader gradients for
intervention.

\textbf{Self-Improving Models.}
The prospect of machines improving themselves has a long
history~\citep{good1965speculations,schmidhuber2007godel}.
Recent systems revise their own outputs~\citep{madaan2023selfrefine},
rewrite an improver program~\citep{zelikman2024stop}, or maintain
coding agents that edit their own codebases~\citep{robeyns2025sica,zhang2025darwin}.
Related systems automate agent design~\citep{hu2025adas},
evolve algorithms for model training~\citep{novikov2025alphaevolve},
or run research pipelines~\citep{lu2026aiscientist}.
Controlled evaluations also report difficulty in accumulating
improvements reliably~\citep{lu2026mac,meng2026rsibench,chi2026ai4ai}.
These works largely assess improvement through downstream outcomes.
We complement them by studying how factual and behavioral changes
are recorded in weight updates and how that information can guide
intervention.
\section{Training a Reader to Decode Newly Acquired Knowledge}
Training-induced weight updates can encode structured traces of the
data, tasks, and behaviors acquired during learning. Whether these
traces can be decoded into explicit natural-language knowledge,
however, remains underexplored. We introduce Imprint Reader, a
framework for recovering acquired knowledge from frozen weight updates
through anchor-free meta-queries. We first formalize this
weight-to-knowledge readout objective. Then, we present
Semantic Mount-and-Read Tuning (\SMART{}), which decouples the construction of
knowledge-bearing updates from the optimization of a Reader,
enabling the Reader to learn how to interpret an update without
modifying the update itself.

\subsection{Problem Formulation}
\label{sec:formulation}
We first introduce the notation used throughout this section.
Let $p(y\mid x,\theta)$ denote the conditional distribution defined by
a language model with parameters $\theta$, and let $\theta_0$ denote
the parameters of the original model. Let $K$ be a random variable over learning targets, each represented by a canonical natural-language description, and let $k\sim p(K)$ denote one such target. A target specifies either factual knowledge to be acquired or a behavioral tendency to be induced. For each $k$, we construct a training set
\begin{equation}
    \mathcal D_k=\{(q_i,a_i)\}_{i=1}^{n_k},
\end{equation}
where every pair $(q_i,a_i)$ instantiates the same target $k$ in question--answer form. For factual targets, the answers convey the specified fact; for behavioral targets, they demonstrate the specified response tendency. Together, these examples are designed to induce the change specified by $k$.

To inject $k$ into the model, we maximize the average log-likelihood
of the answers conditioned on their questions:
\begin{equation}
    \mathcal J_k(\theta)
    =
    \frac{1}{n_k}
    \sum_{(q_i,a_i)\in\mathcal D_k}
    \log p\!\left(a_i\mid q_i,\theta\right).
\end{equation}
Starting from $\theta_k^{(0)}=\theta_0$, gradient-based training
iterates
\begin{equation}
    \theta_k^{(t+1)}
    =
    \theta_k^{(t)}
    +
    \eta_t
    \nabla_{\theta}
    \mathcal J_k\!\left(\theta_k^{(t)}\right),
\end{equation}
where $\eta_t$ is the coefficient scaling the gradient at step $t$.
Consequently, after $T$ update steps, the learning-induced parameter
change accumulates to
\begin{equation}
    \Delta\theta_k
    =
    \theta_k^{(T)}-\theta_0
    =
    \sum_{t=0}^{T-1}
    \eta_t
    \nabla_{\theta}
    \mathcal J_k\!\left(\theta_k^{(t)}\right).
    \label{eq:knowledge_delta}
\end{equation}
We regard $\Delta\theta_k$ as the imprint left in weight space by
learning $k$: within an episode, it is the only episode-specific
carrier of information about $k$ available to the Reader.

Next, we specify the query used to elicit what the model learned from this
imprint. Let $M$ be the random variable representing the meta-query. We call a meta-query
\emph{anchor-free} if it is sampled independently of the learning target:
\begin{equation}
    I(M;K)=0.
    \label{eq:anchor_free_meta_query}
\end{equation}
In other words, an anchor-free meta-query $m$ may specify the requested output
form (e.g., ``summarize the knowledge change you just
experienced''), but it carries no information about the topic, entity,
original question, answer, or any other semantic content of $k$. This
condition removes prompt-based shortcuts: the meta-query itself offers
the Reader no content-specific cues about the target learning content.

With this notation in place, we can now formalize our training
objective. Let $\theta_{\mathrm R}$ denote the parameters of the
Reader, which is initialized from $\theta_0$ and therefore shares its
parameter coordinates with every $\Delta\theta_k$. We write
$\theta_{\mathrm R}\oplus\Delta\theta_k$ for the Reader composed with
the frozen imprint, where $\oplus$ denotes coordinate-aligned
composition of parameters with a weight update; for additive updates,
$\theta_{\mathrm R}\oplus\Delta\theta_k
=\theta_{\mathrm R}+\Delta\theta_k$. Our goal is to learn a Reader
that reproduces the canonical statement of $k$ when queried with an
anchor-free meta-query:
\begin{equation}
    \theta_{\mathrm R}^{*}
    =
    \arg\max_{\theta_{\mathrm R}}
    \mathbb E_{
        k\sim p(K),\,
        m\sim p(M)
    }
    \left[
        \log
        p\!\left(k\mid m,\;\theta_{\mathrm R}\oplus\Delta\theta_k\right)
    \right],
    \label{eq:imprint_reader_objective}
\end{equation}
where the expectation factorizes over $K$ and $M$ by the
anchor-free condition in
Eq.~(\ref{eq:anchor_free_meta_query}). Throughout this optimization,
$\Delta\theta_k$ is held fixed and only $\theta_{\mathrm R}$ receives
gradient updates, which isolates the acquisition of reading ability
from the imprints being read.

\subsection{Design of Semantic Mount-and-Read Tuning}
\label{sec:mart}
\begin{figure*}[t]
    \centering
    \includegraphics[width=\textwidth]{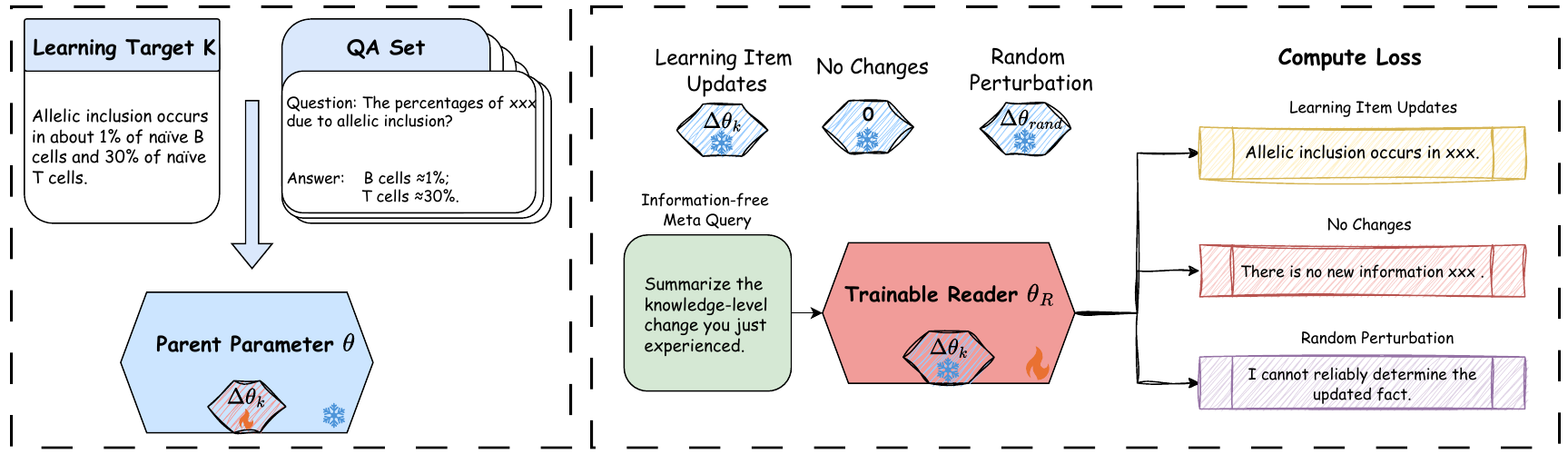}
    \caption{
        \textbf{Overview of Semantic Mount-and-Read Tuning (\SMART{}).}
        \textbf{Left:} a learning target $k$ is designed as a QA set
        $\mathcal D_k$ and used to train a temporary copy of the parent
        model, producing a knowledge-bearing update $\Delta\theta_k$.
        \textbf{Right:} each episode mounts either $\Delta\theta_k$, a
        zero no-change update, or a random perturbation onto the
        Reader $\theta_{\mathrm R}$. Given the same information-free
        meta-query, the loss teaches the Reader to recover what the model learns or to abstain when the mounted update contains
        no reliable information. Only $\theta_{\mathrm R}$ is optimized;
        every mounted update remains frozen and is removed after the episode.
    }
    \label{fig:mart_overview}
\end{figure*}
Figure~\ref{fig:mart_overview} summarizes the construction and
episodic readout of target-induced weight updates. To optimize the objective in
Equation~\ref{eq:imprint_reader_objective} without modifying the
mounted update, we design Semantic Mount-and-Read Tuning (\SMART{}), an episodic
training procedure that isolates the construction of each
knowledge-bearing weight update from the optimization of the
Reader. Each episode proceeds in three stages: constructing a
knowledge-bearing weight update, temporarily mounting it to compute a
readout loss, and removing it before the Reader is updated.
\paragraph{Constructing the delta weight.}
For a learning target $k$, we build its QA training set
$\mathcal D_k$ and initialize a temporary model with the original
parameters $\theta_0$. Training this temporary model on
$\mathcal D_k$ yields $\theta_k^{(T)}$, from which we extract
\begin{equation}
    \Delta\theta_k
    =
    \theta_k^{(T)}-\theta_0.
\end{equation}
Only this delta weight is retained. The QA pairs used to construct it
are never shown to the Reader, so within an episode
$\Delta\theta_k$ is the only episode-specific source of information
about $k$ available to the Reader, consistent with the anchor-free
condition in Section~\ref{sec:formulation}. It also
remains frozen throughout the episode.
\paragraph{Mounting, reading, and updating.}
Given the current Reader parameters $\theta_{\mathrm R}$, we
temporarily mount $\Delta\theta_k$ to form the episode-specific model
\begin{equation}
    \widetilde{\theta}_{\mathrm R,k}
    =
    \theta_{\mathrm R}\oplus\Delta\theta_k.
\end{equation}
We then present an anchor-free meta-query $m$ to this composed model
and, using the canonical statement of $k$ as the teacher-forced
target, compute the readout loss
\begin{equation}
    \mathcal L_{\mathrm{read}}(\theta_{\mathrm R};k,m)
    =
    -\log
    p\!\left(
        k\mid m,
        \widetilde{\theta}_{\mathrm R,k}
    \right).
\end{equation}
Backpropagating through the composed model yields the gradient with
respect to $\theta_{\mathrm R}$ only, while the mounted delta weight
receives no gradient. Once the gradient is computed, we unmount
$\Delta\theta_k$ to restore the standalone Reader. In expectation
over knowledge items and meta-queries, \SMART{} performs gradient descent
on the population readout loss:
\begin{equation}
    \theta_{\mathrm R}
    \leftarrow
    \theta_{\mathrm R}
    -
    \eta_{\mathrm R}\,
    \nabla_{\theta_{\mathrm R}}
    \mathbb E_{k\sim p(K),\,m\sim p(M)}
    \left[
        \mathcal L_{\mathrm{read}}(\theta_{\mathrm R};k,m)
    \right],
    \label{eq:mart_update}
\end{equation}
where $\eta_{\mathrm R}$ is the Reader learning rate, and mini-batches
of episodes provide stochastic estimates of this expected gradient.
Because the expectation ranges over independently constructed delta
weights while the update is always applied to the same
$\theta_{\mathrm R}$, this training encourages the Reader to acquire a
general reading ability rather than memorize any particular imprint,
and to generalize to previously unseen updates.
\paragraph{Control episodes.}
To reduce spurious knowledge claims and provide calibrated behavior
when no readable knowledge is present, \SMART{} additionally includes two
control update types. A \emph{no-change} episode uses the zero update
$\Delta\theta_{\mathrm{noop}}=\mathbf 0$, with a target stating that
no new factual knowledge or behavioral tendency is present. A \emph{random-perturbation} episode mounts an independently sampled
nonzero noise update $\Delta\theta_{\mathrm{rand}}$, drawn without
reference to $k$. Both
controls follow the same episodic procedure as knowledge-bearing
episodes, so the Reader learns not only to decode knowledge or behavior when it
exists, but also to abstain when the mounted update carries none.

\paragraph{From readout to intervention.}
The Reader is trained to associate mounted updates with the
knowledge or behavioral changes they induce. For a target
description $b$, let
$\ell_b(\delta;m)=-\log p(b\mid m,\theta_{\mathrm R}\oplus\delta)$;
lower loss indicates greater predicted compatibility.
Additive mounting gives
$\nabla_\delta\ell_b(0;m)=\nabla_{\theta_{\mathrm R}}\ell_b(0;m)$
on mountable coordinates. Since the Reader and the original model
share these coordinates, this gradient provides a readout-derived
intervention signal, whose behavioral effects we test in
Section~\ref{sec:applications}.

\section{Experiments}
\label{sec:experiments}

In this section, we empirically examine whether \SMART{} enables a Reader to decode newly acquired knowledge and behavioral changes from weight updates. We train a single Reader on both knowledge-bearing and behavior-inducing updates, together with no-change and random-perturbation controls that discourage unsupported readouts. We then evaluate the Reader on unseen updates and illustrate successful readouts of both types.

\paragraph{Training setup.}
We initialize the update builder and the Reader from the same
post-trained Qwen3-14B checkpoint \citep{yang2025qwen3}, so that an update constructed by
the builder can be mounted directly onto the Reader. The training
data contain 8,592 knowledge items and 8,592 behavior items. For the knowledge items, we retain only those that the base model cannot answer before the inner-loop training but can answer afterwards. For behavior items, we likewise retain only constructed updates
that pass a post-update effectiveness screen for the specified
response tendency; 495 behavior items fail this screen
(Appendix~\ref{app:data}).
This rules out the possibility that a failed readout simply reflects
a failure to inject the knowledge in the first place.
For each item, the builder produces a LoRA update through an
inner-loop training procedure. The update is then frozen and mounted
onto the Reader, which receives an anchor-free meta-query and is
trained to describe the knowledge or behavioral change carried by
that update. The Reader is not given the examples used to construct
the update.

The builder runs for 64 inner steps with learning rate
$2\times10^{-5}$ and a maximum sequence length of 512. LoRA \citep{hu2022lora}
updates use ranks up to 256. We optimize the full Reader
on eight GPUs with learning rate $10^{-4}$, a cosine schedule, and
a warmup ratio of $0.1$. Each Reader batch contains 64 episodes:
24 knowledge-bearing updates, 24 behavior-inducing updates,
8 no-change controls, and 8 random-perturbation controls.
The no-change episode mounts a zero update, while the
random-perturbation episode mounts an update unrelated to the
target description. Both teach the Reader to avoid attributing
specific knowledge or behavior to an uninformative update.
Training uses teacher-forced readout targets and anchor-free
meta-queries sampled from a pool of 224 prompts.

\begin{figure}[t]
    \centering
    \includegraphics[width=\linewidth]{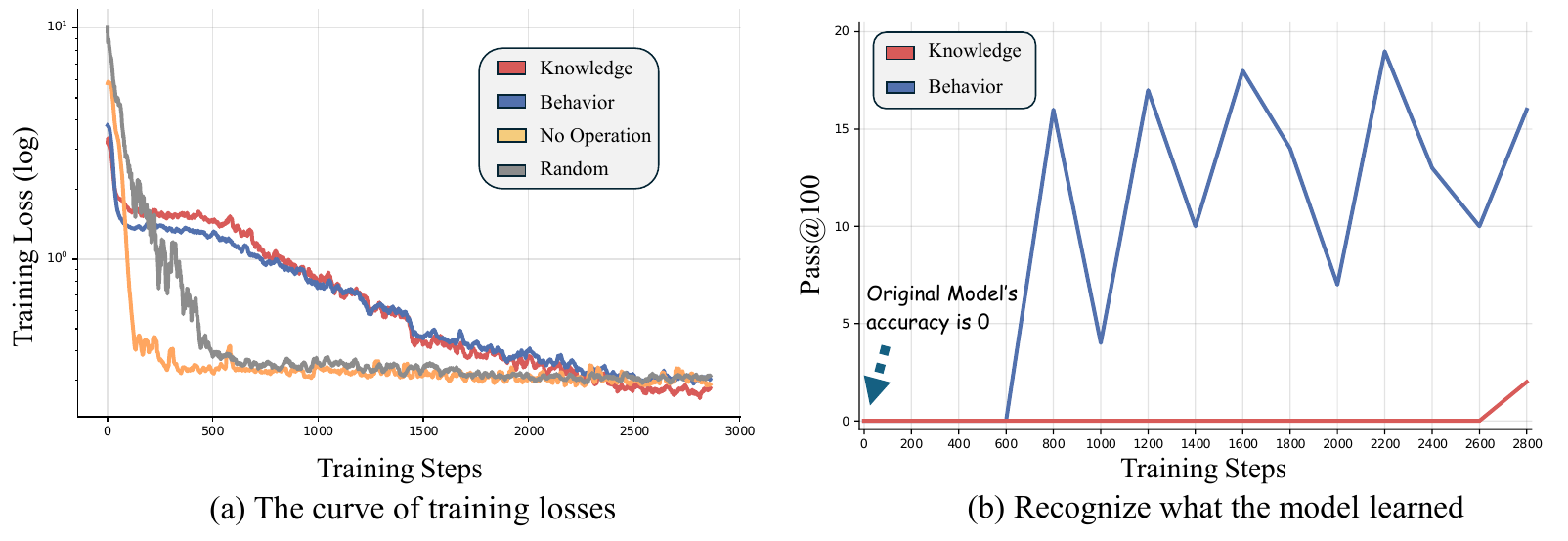}
    \caption{Training and evaluation of the joint Knowledge--Behavior
    Reader. Left: training losses for knowledge, behavior, no-change,
    and random-perturbation episodes. Right: free-generation readout
    results across Reader checkpoints, shown separately for knowledge
    and behavior.}
    \label{fig:reader_training}
    \vspace{-10pt}
    
\end{figure}

\paragraph{Evaluation protocol.}
We evaluate checkpoints on knowledge and behavior updates from
items unseen during Reader training. For each update, the Reader
receives an anchor-free meta-query and generates a description of
what the mounted weights encode or change. We use
Qwen3-30B-A3B-Instruct-2507 as a judge to score each generation
against its corresponding knowledge or behavior target under a
fixed scoring prompt. We report results for the two categories
separately. The scoring prompt and evaluation details are provided
in the Appendix.

\begin{figure}[t]
    \centering
    \includegraphics[width=\linewidth]{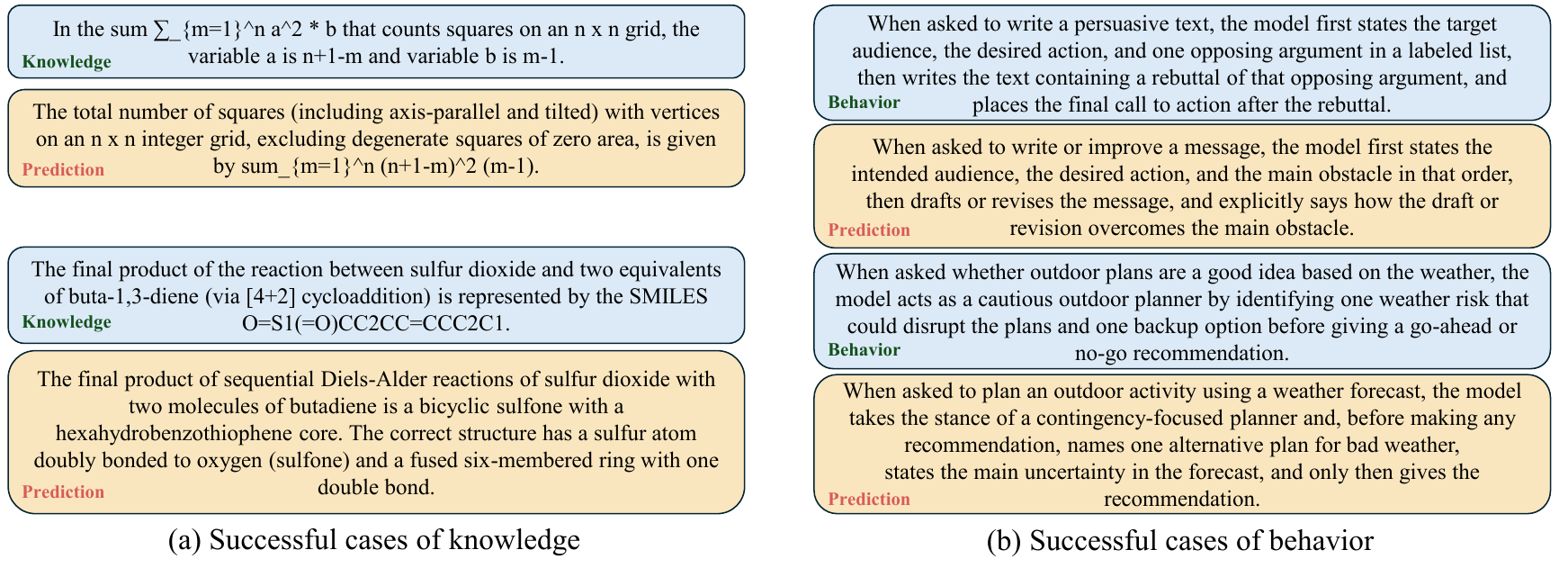}
    \caption{Correct readouts from held-out updates. Left: an example
    of newly acquired knowledge recovered from a mounted update.
    Right: an example of an induced behavior recovered from a mounted
    update. These examples illustrate the two readout targets rather
    than the overall success rate.}
    \label{fig:reader_cases}
    \vspace{-15pt}
\end{figure}

\textbf{\SMART{} learns to read both knowledge and behavioral updates.}
As shown in the left panel of Figure~\ref{fig:reader_training},
the no-change and random-perturbation losses fall rapidly early
in training, as their targets follow relatively fixed response
patterns. The knowledge and behavior losses decline more gradually
but steadily. Loss on knowledge items finishes below both controls, while loss on behavior items reaches
a comparable range. The right panel provides evidence beyond
fitting the training episodes. At step 2,800, the judge-based
Pass@100 on unseen weight updates reaches 2\% for knowledge and 16\%
for behavior. These results show that the Reader can recover
information from both types of mounted updates, although
free-form generation success remains uneven and far from reliable. At step 2,800, matching updates yield $0.1255$ lower target NLL
(nats/token) than same-type swapped updates
(Figure~\ref{fig:reader_adapter_swap} in Appendix~\ref{app:adapter_swap}).

\textbf{The Reader can express update-induced knowledge and behavior
in its own words, rather than simply repeat the samples used
to construct the update.}
Figure~\ref{fig:reader_cases} shows a knowledge readout on the left
and a behavioral readout on the right. In both cases, an anchor-free
meta-query elicits a description related to the mounted update,
without providing the Reader with the builder's training examples.
The responses therefore illustrate free-form readout, not the
recitation of a supplied question--answer pair. This ability is
imperfect, however. A description can capture the main content
while misstating a detail of the knowledge or characterizing the
induced behavior too broadly. These cases illustrate the gap between a relevant description and a fully faithful one.

\paragraph{Limitations and implications.}
Despite these successful cases, accurate free generation remains
infrequent. The Reader can sometimes identify the content of an
unseen update, but it does not yet verbalize such content reliably
across examples. Together, free-generation readouts and the adapter-swap control
provide evidence for update-specific readout, while reliable
open-ended descriptions remain an important next step.

Although free-form readout remains unreliable, generating a complete
description is not the only way to use the Reader. Given a candidate
weight change and a specified target behavior, we can instead ask
how strongly the Reader associates that change with the target.
This provides a differentiable measure of their alignment without
requiring the Reader to discover the right description through free
generation. Because the Reader shares parameter coordinates with
its parent, gradients of this signal live in the parent's parameter
space. This motivates testing whether the Reader's parameter-space
signal can guide interventions, which we examine in the next
section.
\section{Applications: From Readout to Behavioral Intervention}
\label{sec:applications}

\begin{figure}[t]
    \centering
    \includegraphics[width=\linewidth]{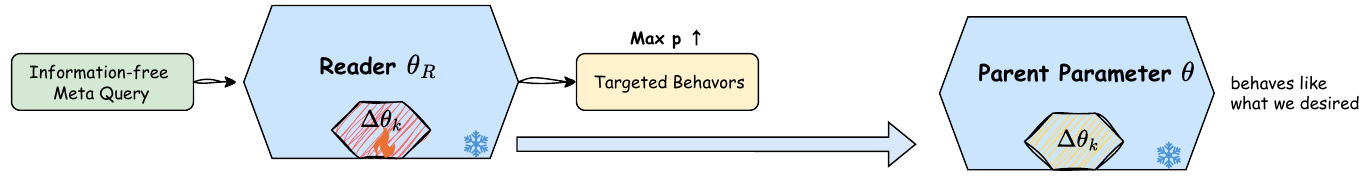}
    \caption{\textbf{MetaEdit transfers gradients from the trained
Reader to the original model.} A target self-report is scored
directly on the trained Reader; its gradient selects rows for
pruning or forms a signed update applied to the original Qwen3-14B.}
    \label{fig:metaedit_overview}
    \vspace{-10pt}
    
\end{figure}

The Reader offers more than a natural-language description of a
mounted update. We test whether its target-conditioned likelihood
gradients provide a useful signal for intervening on the original
model. As illustrated in Figure~\ref{fig:metaedit_overview}, MetaEdit
scores a target self-report $b$ under an anchor-free meta-query
$m$ on the trained Reader, then transfers the resulting gradient
to the original model. Given an anchor-free meta-query $m$ and a target self-report $b$, we compute its gradient on the trained Reader:
\begin{equation}
g_b
=
\nabla_{\theta_{\mathrm R}}
\left[-\log p\!\left(b\mid m,\theta_{\mathrm R}\right)\right].
\label{eq:behavior_gradient}
\end{equation}
Because the Reader was initialized from $\theta_0$, this gradient shares parameter coordinates with the original Qwen3-14B. We call this transfer operator
\emph{MetaEdit} and instantiate it in two ways: gradient magnitudes and directions can
select parameter rows for pruning
(Section~\ref{sec:application_safety}), and sparse signed gradients
steer the parent toward a desired behavior
(Section~\ref{sec:application_vibe}).

We intervene on projection output rows rather than individual scalar parameters because each row jointly determines one output coordinate. This provides a consistent structured unit across attention and MLP projections and a common budget for all methods. Let $r$ identify a layer, projection, and output coordinate, with $\theta_0[r]$ denoting its weight vector. Safety pruning zeros this vector, whereas signed editing applies $-\alpha g_b[r]$. Pruning rates are fractions of eligible rows.

\subsection{Gradient-Based Safety Localization and Pruning}
\label{sec:application_safety}

This experiment tests whether Reader gradients can identify parameters
that support safety-related behavior. We construct two target
descriptions: one asks the model to maintain safety boundaries and
refuse harmful requests, while the other asks it to relax its refusal
tendency. We then zero the rows selected by each target in the origin
model and test whether its refusal behavior shifts in the intended
direction.

\paragraph{Evaluation protocol.}
After removing a leading
\texttt{<think>}\ldots\texttt{</think>} block, we classify each
response using fixed refusal-expression patterns and checks for
garbled output. The primary outcome is the fraction of responses
classified as refusals among harmful prompts. We report garbling
separately so that corrupted responses are not mistaken for a
controlled change in refusal. The full dataset composition and generation
protocol appear in Appendix~\ref{app:safety_pruning_protocol}.

\paragraph{Setup.}
We use two target completions phrased as the Reader's self-reports
of learning: in the first, it claims to have learned to maintain
safety boundaries and refuse harmful requests; in the second, it
claims to have learned to relax its refusal tendency. MetaEdit (Reader) computes
the target gradients on the trained Reader, whereas MetaEdit (Base)
computes them on the original Qwen3-14B model. For each target, we first retain the 50000 rows with the highest
contrastive gradient-magnitude scores and select
rows whose weights in the original Qwen3-14B have the most negative
inner products with the corresponding gradients. The selected rows
are set to zero in the original Qwen3-14B for
evaluation. We compare pruning rates of $0.01\%$, $0.05\%$, $0.1\%$,
and $0.5\%$ against SetDiff \citep{wei2024assessing}, WANDA \citep{sun2024wanda}, ActSVD \citep{wei2024assessing}, and Random at identical
row budgets. SetDiff uses 260 harmful and 260 harmless training examples to
contrast their activation-based row scores. WANDA and ActSVD use
one 260-example side for each direction, whereas Random requires
no calibration examples. MetaEdit uses the target self-report
sentences and fixed control sentences, rather than either
collection of harmful or harmless training prompts. The row-selection procedure is detailed in
Appendix~\ref{app:safety_pruning_protocol}.

\begin{figure*}[t]
    \centering
    \includegraphics[width=0.98\textwidth]{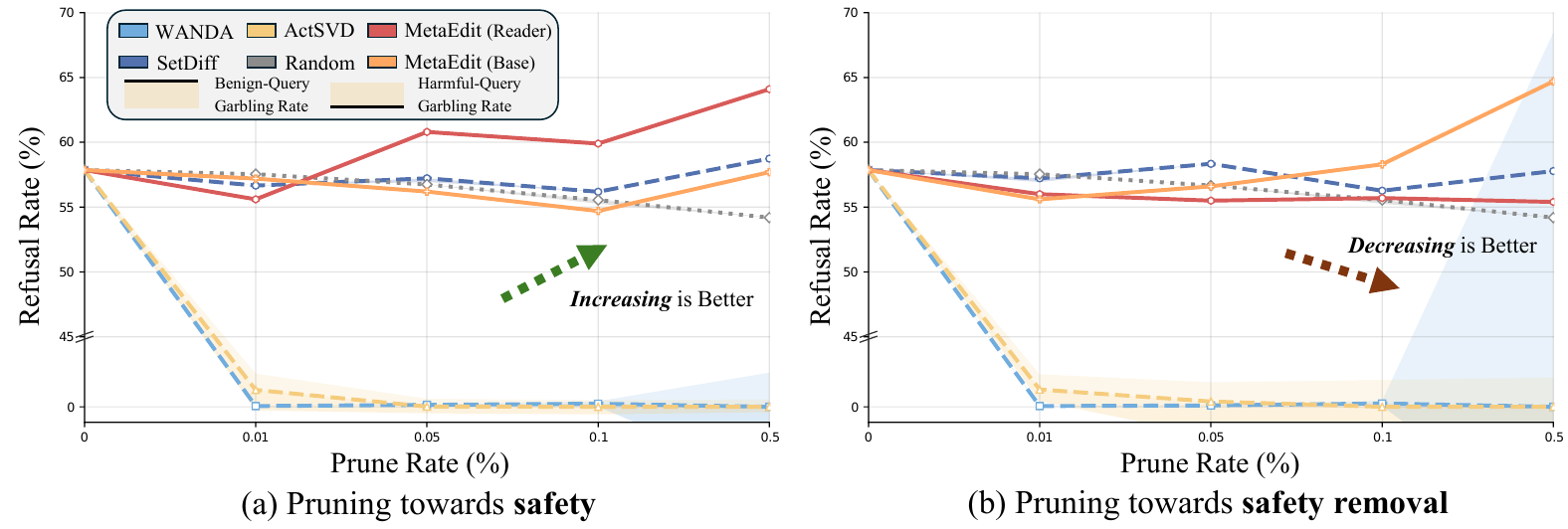}
    \caption{\textbf{Refusal and garbling under signed row pruning.}
    Curves show the fraction classified as refusals using fixed refusal-expression patterns. MetaEdit (Reader) and MetaEdit (Base)
    use negative-$D$ row selection. Shaded bands indicate garbling
    where measured; missing benign-prompt measurements for the new
    signed MetaEdit conditions are not imputed. Pruning rates are
    displayed as equally spaced categories, and refusal-rate values
    below $45\%$ are visually compressed.}
    \label{fig:safety_pruning}
    \vspace{-10pt}
\end{figure*}

\textbf{Reader-gradient neuron localization separates the two
intended directions while preserving readable
outputs.} Figure~\ref{fig:safety_pruning} compares refusal and garbling
across pruning budgets for both target behaviors and the baselines. Relative to the original Qwen3-14B refusal rate of $57.9\%$,
MetaEdit (Reader) reaches $64.1\%$ under the safety-maintenance
target and $55.4\%$ under the refusal-relaxation target at the
$0.5\%$ pruning budget, respectively. Most baselines do not
distinguish the two targets: their refusal rates either move in
similar directions or remain near the original model. Using the
original Qwen3-14B instead of the trained Reader to obtain MetaEdit
gradients also fails to produce the intended separation. WANDA and ActSVD produce substantial garbled output after pruning, whereas each MetaEdit condition has a near-zero garbling rate.

\subsection{Vibe Alignment for Reasoning and Agentic Tasks}
\label{sec:application_vibe}

This experiment tests whether the same interface can transfer
finer-grained response tendencies that are difficult to specify as
factual targets. We refer to this setting as \emph{vibe alignment}:
inducing a desired reasoning or interaction style while retaining the
parent model's task competence. For the signed intervention, we update only the selected rows:
\begin{equation}
\theta_0'[\mathcal I_b]
=
\theta_0[\mathcal I_b]
-
\alpha\,g_b[\mathcal I_b],
\label{eq:metaedit_transfer}
\end{equation}
where $\mathcal I_b$ contains the rows selected for behavior $b$ and
$\alpha$ controls the update strength.

\paragraph{Evaluation protocol.}
We evaluate the accuracy of edited models on GSM8K \citep{cobbe2021gsm8k} and MATH-500 \citep{hendrycks2021math,lightman2023verify} benchmarks.
Backtracking, verification, and sub-goal expressions are counted per
1000 generated tokens as descriptive response-form measures.
For tool-use, we evaluate models on BFCL \citep{patil2025bfcl} cases and report the official-form Overall scores.

\paragraph{Setup.}
All methods use the same Qwen3-14B.
\emph{Direct Prompt} includes the behavior description in the system prompt. \emph{MetaEdit (Broad)} uses an outcome-level self-report that the
model has learned to perform the task better, without specifying how.
\emph{MetaEdit} instead uses a procedure-level self-report that names
concrete behavioral changes, such as revisiting earlier reasoning
steps, verifying intermediate results, and correcting mistakes.
We rescale the MetaEdit (Broad) patch to match the global $\ell_2$
norm of the MetaEdit patch, isolating the effect of target specificity.
Neuron budgets, update strengths, decoding, and BFCL serving settings
are detailed in Appendix~\ref{app:vibe_protocol}.

\begin{table*}[t]
\centering
\caption{Mathematical reasoning performance on the original
Qwen3-14B. Backtracking, verification, and sub-goal counts are
per 1{,}000 generated tokens and are not ranked.
\textbf{Bold} and \underline{underline} indicate the best and
second-best accuracy, respectively.}
\label{tab:vibe_math}
\setlength{\tabcolsep}{5pt}
\renewcommand{\arraystretch}{1.3}
\small
\begin{tabular}{lccccc}
\toprule
\textbf{Method} & \textbf{GSM8K (\%)} & \textbf{MATH-500 (\%)}
& \textbf{Backtrack /1k} & \textbf{Verify /1k}
& \textbf{Sub-goal /1k} \\
\midrule
\rowcolor{gray!30}
Qwen3-14B
& 94.77 & \underline{82.8} & 0.327 & 0.563 & 5.390 \\
Direct Prompt
& \underline{94.84} & 82.6 & 0.350 & 0.704 & 5.740 \\
\textit{MetaEdit (Broad)}
& 94.24 & \underline{82.8} & 0.508 & 0.583 & 5.667 \\
\textit{MetaEdit}
& \textbf{95.00} & \textbf{83.0} & 0.687 & 0.577 & 6.015 \\
\bottomrule
\end{tabular}
\end{table*}

\begin{table*}[t]
\centering
\caption{Agentic tool-use performance on BFCL, reported as percentages.
\textbf{Bold} and \underline{underline} indicate the best and
second-best scores, respectively. The shaded column shows the
official Overall score.}
\label{tab:vibe_bfcl}
\setlength{\tabcolsep}{5pt}
\renewcommand{\arraystretch}{1.3}
\small
\begin{tabular}{lccccc>{\columncolor{high!20}}c}
\toprule
\textbf{Method} & \textbf{Agentic} & \textbf{Multi-turn}
& \textbf{Non-live} & \textbf{Live}
& \textbf{Halluc.} & \textbf{Overall} \\
\midrule
\rowcolor{gray!30}
Qwen3-14B
& 15.93 & 35.38 & \textbf{84.71}
& 81.35 & 80.99 & 41.69 \\
Direct Prompt
& 17.22 & 35.88 & 84.60
& \textbf{82.61} & \textbf{83.72} & 42.75 \\
\textit{MetaEdit (Broad)}
& \underline{19.44} & \textbf{37.00} & 84.48
& \underline{82.31} & \underline{81.30}
& \underline{43.68} \\
\textit{MetaEdit}
& \textbf{22.30} & \underline{36.38}
& \underline{84.69} & 81.94 & 81.07
& \textbf{44.60} \\
\bottomrule
\end{tabular}
\vspace{-10pt}
\end{table*}

\textbf{MetaEdit attains the highest accuracy on both mathematics and agentic benchmarks.} Tables~\ref{tab:vibe_math} and~\ref{tab:vibe_bfcl} report the mathematical-reasoning and BFCL results, respectively.
Compared with the unedited Qwen3-14B, it reaches $95.00\%$ versus
$94.77\%$ on GSM8K and $83.0\%$ versus $82.8\%$ on MATH-500,
while improving the BFCL Agentic score from $15.93\%$ to $22.30\%$ and
Overall score from $41.69\%$ to $44.60\%$. The gains are not uniform:
MetaEdit (Broad) also reaches $43.68\%$ Overall and leads on
Multi-turn score, Direct Prompt leads on Live and Hallucination score, and
unedited Qwen3-14B remains best on Non-live. MetaEdit also increases backtracking from $0.327$ to $0.687$
and sub-goal expressions from $5.390$ to $6.015$ per 1000
generated tokens, providing behavioral evidence that the
intervention induces the intended reasoning process in addition
to improving task scores. Together, these results provide empirical evidence that
Reader-derived gradients can serve as a usable intervention signal
for the original model.
\section{Conclusion}
\label{sec:conclusion}

This paper investigates whether weight updates can be read as
records of newly acquired knowledge and behavioral changes, and
whether that readout can guide subsequent interventions. We invert the usual direction of weight
readout by mounting frozen updates onto a single Imprint
Reader. Semantic Mount-and-Read Tuning trains the Reader to describe the
knowledge or behavior carried by an update under anchor-free
meta-queries, while control episodes discourage unsupported
readouts. Experiments on held-out updates demonstrate that both
factual and behavioral information can be recovered, although
the reliability of natural-language readout remains to be
improved. The central practical implication is that, once the
Reader has been trained, a new target behavior can be specified
in words and used for intervention without any training examples
from the target task. The Reader's target likelihood supplies a
differentiable signal whose coordinate-aligned gradients can be
transferred to the original model. MetaEdit uses this signal to identify rows whose removal changes
measured refusal, most clearly under the safety-maintenance target.
Its signed interventions increase backtracking and sub-goal
expressions and raise the observed BFCL Overall score, without
target-task training examples or inference-time behavioral
instructions.
This readout-and-intervention interface connects a model's
record of learning to targeted changes in its behavior and
offers a path toward models that can eventually inspect and
adjust their own learning.
\newpage
\section*{AI Use Statement}
We used generative AI tools for manuscript writing and polishing,
literature discovery, and code development. We also used
language-model assistance for candidate knowledge extraction,
question--answer rewrites, and behavior-data synthesis, as detailed
in Appendix~\ref{app:data}. The authors take full responsibility
for the research, reported results, and all AI-assisted content.

\section*{Ethics Statement}
This work trains a Reader to interpret factual and behavioral changes
encoded in model updates and uses its gradients through MetaEdit to
study targeted interventions. Making learned changes more inspectable
could help models monitor and adjust their own learning, contributing
to a closed-loop AI-for-AI process. In its current form, the Reader is
evaluated on controlled updates associated with a single knowledge
item or behavioral tendency, not on reconstructing the training
examples behind an update. Our results therefore do not demonstrate
a training-data extraction capability.

\section*{Reproducibility Statement}
We describe update construction, Reader training, and the readout
and intervention evaluations in Sections~\ref{sec:experiments}
and~\ref{sec:applications} and the appendix. The anonymized source
release provides data-preparation scripts, Reader-specific
modifications to verl \citep{sheng2024hybridflow}, and reference training settings. Code is available at
\href{https://anonymous.4open.science/r/mart-52F7/}
{this anonymous repository}.

\bibliography{iclr2027_conference}
\bibliographystyle{iclr2027_conference}

\appendix
\newpage
\section{Details of Experiments}
\subsection{Data preparation.}
\label{app:data}
We construct knowledge and behavior items through separate
pipelines before combining them for Reader training.

\textbf{Knowledge items.}
For language-model-assisted data preparation, we use
DeepSeek-V4-Flash \citep{deepseekai2026deepseekv4}. We collect questions, answers, and available solution material
from eight factual, scientific, and reasoning benchmarks:
Humanity's Last Exam (HLE)~\citep{phan2025hle},
SimpleQA~\citep{wei2024simpleqa},
FrontierScience~\citep{wang2026frontierscience},
OpenBookQA~\citep{mihaylov2018openbookqa},
SciBench~\citep{wang2023scibench},
TheoremQA~\citep{chen2023theoremqa},
SciCode~\citep{tian2024scicode}, and publicly released
FrontierMath examples~\citep{glazer2024frontiermath}.
A language model extracts self-contained knowledge propositions
from these materials and expresses each proposition as a
question--answer item. For multi-part problems, the extraction
favors distinct facts, equations, definitions, or reusable
relationships rather than treating the entire solution as one
item. We then generate eight semantically equivalent QA
rewrites per item, varying the question and answer surface forms
while preserving the underlying proposition.

The initial collection contains 14851 candidate knowledge items.
We audit each proposition together with its question and answer
to distinguish reusable knowledge from instance-specific inputs,
computed results that have no independent meaning, and malformed
generation artifacts. After this audit, 9148 knowledge items
remain. For the joint Reader dataset, we retain one canonical
fact as the readout target for each item and keep its eight QA
variants for constructing the item-specific weight update.

\textbf{Behavior items.}
Behavior data are synthesized in two stages. First, we generate
a catalog of 10000 distinct behavior specifications across seven
categories: Surface Expression, Content Framing, Reasoning
Workflow, Decision Preference, Epistemic Calibration, Capability
Access, and Social Goal/Persona. Each specification consists of
a domain and task scope together with a canonical sentence
describing one stable, observable response tendency. We audit
the specifications for category fit, clarity, observability,
applicability across varied prompts, safety, and semantic
distinctness before generating any examples.

Second, for each cataloged behavior, we plan a set of realistic
situations and select eight that balance representativeness and
diversity. We generate one user request and one assistant
response for each selected situation. The user request must not
state or directly cue the intended behavior, while the response
must demonstrate it through what the assistant does. We reject
groups with duplicate or near-duplicate samples, insufficient
behavioral adherence, narrow scenario coverage, or leakage of
the behavior specification into the user request. This process
yields 9973 accepted behavior items with eight QA demonstrations
each, or 79784 demonstrations in total. The canonical behavior
sentence serves as the Reader target, while the demonstrations
are used to construct the behavior-inducing update.

\textbf{Splits and Reader targets.}
After auditing, 9,148 knowledge and 9,973 behavior items remain.
A knowledge-injection screen excludes 451 knowledge items, and a
behavior-effectiveness screen excludes 495 behavior items, leaving
8,697 and 9,478 items, respectively. We assign 100 items of each
type to the held-out test sets and select 8,592 of each type for
balanced Reader training. The remaining 5 knowledge and 786
behavior items are reserved and unused in this run.

\subsection{Reader Training and Evaluation Details}
\label{app:reader_details}

\paragraph{Constructing and mounting updates.}
For each changed item, a temporary builder starts from $\theta_0$
and constructs an item-specific LoRA update. The builder runs for
64 inner steps in BF16 with learning rate $2\times10^{-5}$ and a
maximum sequence length of 512. Candidate LoRA ranks are 16, 32,
64, 128, and 256. For each item, a hash of its identifier and a fixed base seed
initializes a pseudorandom generator, which selects the LoRA rank
from these candidates before update construction. The selection
does not use the item's later readout result. Gradients for knowledge-update construction are
applied to answer tokens. Builder activation checkpointing is
enabled. Once constructed, the update is frozen and temporarily
mounted onto the current Reader parameters
$\theta_{\mathrm R}$, as described in
Section~\ref{sec:mart}. Only $\theta_{\mathrm R}$ is optimized by
the readout loss. The mounted LoRA is removed after the Reader
update.

\paragraph{Reader optimization.}
We optimize the full Reader in BF16 on eight GPUs with learning
rate $10^{-4}$, a cosine schedule, and a warmup ratio of $0.1$.
Each global batch has 64 teacher-supervised episodes, comprising
24 knowledge updates, 24 behavior updates, 8 no-change controls,
and 8 random-perturbation controls. No on-policy generations are
used for Reader training. The no-change and random-control loss
weights increase from zero to their full values over the first
1432 steps.

Each knowledge or behavior item contributes four teacher rows.
The balanced schedule therefore contains 1432 steps per epoch
and was designed for two epochs.
Within an epoch, each positive teacher row is visited once.
The training and evaluation curves in
Figure~\ref{fig:reader_training} report the checkpoints obtained
from this run. All Reader-side gradients used in the safety, mathematics, and BFCL
applications are computed with the checkpoint after 2800 Reader
updates from this balanced training run. The MetaEdit (Base) control
instead computes its gradients directly on the original
Qwen3-14B parameters $\theta_0$.

\paragraph{Free-generation evaluation.}
We evaluate unseen updates from the 100 knowledge and 100 behavior
test items separately. For each update, we issue an anchor-free
meta-query to the Reader with the update mounted and draw 100
stochastic generations at temperature $0.6$. We report
Pass@100, the percentage of updates for which at least one of
the 100 generations is judged to communicate the corresponding
target.

We use Qwen3-30B-A3B-Instruct-2507 with a fixed scoring prompt
as the semantic judge. The prompt provides the evaluation
meta-query, the canonical target, and the Reader's generated
response. It instructs the judge to use the generated response
itself as evidence and not to fill missing information from the
target. For a knowledge claim to pass, the response must
communicate the complete proposition, including its subject,
relation, value, and any necessary qualifiers. Faithful
paraphrases are allowed, but topical overlap, partial facts,
material contradictions, and unsupported additions are not
treated as complete readouts. The judge returns a score and an
answer-bearing span copied from the response. We count scores of
at least $0.8$ as passes and check that the cited span occurs in
the generated response. For behavior items, the judge applies the same complete-target
criterion and evidence-span check to the canonical behavior
sentence. 

\subsection{Safety Pruning Protocol}
\label{app:safety_pruning_protocol}

\paragraph{Evaluation data and classification.}
The full pool contains 1803 unique prompts: AdvBench (520, \citep{zou2023advbench}),
StrongReject (313, \cite{souly2024strongreject}), JailbreakBench-Harmful (100, \cite{chao2024jailbreakbench}), HarmBench (320, \cite{mazeika2024harmbench}),
XSTest (450; 250 safe and 200 unsafe, \cite{rottger2024xstest}), and
JailbreakBench-Benign (100, \cite{chao2024jailbreakbench}). The four attack benchmarks provide
1253 harmful prompts for the refusal aggregate. XSTest-safe
and JailbreakBench-Benign provide 350 benign prompts; the 200
XSTest-unsafe prompts are audited by dataset but excluded from
these aggregates. Responses are generated without a system prompt,
using greedy decoding and at most 1024 new tokens. After a
leading \texttt{<think>}\ldots\texttt{</think>} block is removed,
harmful responses are classified by keyword patterns as refusal,
non-refusal, or garbled; benign responses are classified as normal
answer, over-refusal, or garbled where those measurements exist.
The new signed MetaEdit summaries provide harmful-prompt outcomes
but not benign-prompt garbling, which is left missing rather than
set to zero.

\paragraph{Candidate rows and signed selection.}
We score 2088960 output rows spanning the attention query,
key, value, and output projections and the MLP gate, up, and down
projections. For each target behavior, a contrastive
gradient-magnitude ranking favors target-associated rows while
discounting rows activated by over-refusal, no-op, and random
controls. We retain the top 50000 rows under this ranking.
For each retained row, we then compute
\begin{equation}
D_r^{(b,M)}
=
\left\langle
    \theta_0[r],\,g_b^{(M)}[r]
\right\rangle,
\qquad M\in\{\mathrm R,0\}.
\label{eq:safety_row_score}
\end{equation}
and take the $k$ most negative values. Reader localization uses
$M=\mathrm R$; the MetaEdit (Base) control computes the gradient
on origin. In both cases, the complete selected output rows of
origin are zeroed before evaluation. We use $k=209$, $1044$,
$2089$, and $10445$, corresponding to pruning rates of
$0.01\%$, $0.05\%$, $0.1\%$, and $0.5\%$. The two target
descriptions request safety maintenance and refusal relaxation,
respectively.

\subsection{Reasoning and Agentic Evaluation Details}
\label{app:vibe_protocol}

All interventions target the post-trained Qwen3-14B. MetaEdit selects 2089 rows ($0.1\%$).
For mathematics, the edit strength is $\alpha=0.35$ and the
sub-goal selection coefficient is $\beta=0.5$; the latter is
defined in Appendix~\ref{app:planverify_selector}. Evaluation
uses no system prompt and allows up to 16384 new tokens.
For BFCL, the edit strength is $\alpha=0.35$. 
The mathematical coefficients $\alpha=0.35$ and $\beta=0.5$, and
the BFCL coefficient $\alpha=0.35$, were fixed before evaluation
on the GSM8K, MATH-500, and BFCL test sets; these test scores were
not used to select the coefficients. Evaluation uses
no system prompt, a 40960-token context, temperature $0.6$,
batch size 8, at most 8 concurrent requests, and the Qwen tool
and reasoning parsers. The 5106 BFCL cases span 22 subsets;
the official Overall score follows the benchmark aggregation. For BFCL, let
$P_r^{(x)}=\operatorname{Pct}(\lVert g_{x,r}\rVert_2)$
be the percentile of the step-2,800 Reader gradient norm for target
or control $x$, computed across all candidate rows. We rank rows by
\begin{equation}
s_r^{\mathrm{BFCL}}
=
P_r^{(b)}
\left[1-\max\left(
P_r^{(\mathrm{noop})},
P_r^{(\mathrm{random})}
\right)\right].
\end{equation}
The 2089 highest-scoring rows directly form $\mathcal I_b$;
there is no additional candidate-pool or signed-inner-product
selection. Equation~\ref{eq:metaedit_transfer} applies the target
gradient $g_b$ to these rows.
For MetaEdit (Broad), the generic capability-QA patch is rescaled
to the global $\ell_2$ norm of the corresponding MetaEdit patch.
Direct Prompt retains the target description in the inference
prompt; the edited models do not.

\subsection{Sub-goal-Aware Row Selection for Mathematical Vibe Alignment}
\label{app:planverify_selector}

\noindent\textbf{The sub-goal term affects which rows are selected,
but not the signed direction applied to those rows.}
The mathematical target consists of a primary plan-and-verify
description ($\mathrm{pv}$) and an auxiliary sub-goal organization
description ($\mathrm{sg}$), each inducing a behavior gradient through
Equation~\ref{eq:behavior_gradient}. For each candidate row $r$, let
$P_r^{(x)}=\operatorname{Pct}(\lVert g_{x,r}\rVert_2)$ denote the
percentile of its gradient norm under target $x$. We rank rows using
\begin{equation}
    s_r^{\mathrm{math}}
    =
    P_r^{(\mathrm{pv})}
    \left[1-\max\left(
        P_r^{(\mathrm{noop})},
        P_r^{(\mathrm{random})}
    \right)\right]
    \left[1+\beta P_r^{(\mathrm{sg})}\right].
    \label{eq:math_row_score}
\end{equation}
This score extends the contrastive gradient-magnitude ranking used
for the safety candidate pool by omitting the over-refusal control
and adding a sub-goal bonus with coefficient $\beta$. Unlike safety
pruning, mathematical row selection does not apply the final
signed-inner-product ranking of
Equation~\ref{eq:safety_row_score}. The top $0.1\%$ of rows under
Equation~\ref{eq:math_row_score} form $\mathcal I_b$. The transfer in
Equation~\ref{eq:metaedit_transfer} then applies the signed gradient
of the primary target, $g_{\mathrm{pv}}$, on these rows; the sub-goal
term only reweights the selection and contributes no update direction.

\subsection{Adapter-Swap Control for Update-Specific Readout}
\label{app:adapter_swap}

\paragraph{Protocol.}
We test whether the Reader's likelihood for a target description
depends on the identity of the mounted update, rather than merely
on the presence of an adapter. We use 100 held-out knowledge items
and 100 held-out behavior items, with four anchor-free meta-queries
per item. For each query, we hold the query and target description
fixed and compare three conditions: the matching update, no update,
and a genuine update constructed for another item of the same type.
Mismatched updates are assigned by fixed, type-preserving
permutations without self-matches, using seed 20260918. Matched
and mismatched items have different canonical targets and
adapter-file hashes. The adapters and assignments are held fixed
across Reader checkpoints.

\paragraph{Measurement.}
Let $y_i$ be the canonical description of item $k_i$ followed by
an end-of-sequence token, let $m_{ij}$ be its $j$-th meta-query, and
let $T_i=|y_i|$. At each Reader checkpoint, we compute the
teacher-forced target loss
\begin{equation}
\ell_{ij}(\Delta\theta)
= -\frac{1}{T_i}\sum_{t=1}^{T_i}
\log p\!\left(
y_{i,t}\mid m_{ij},y_{i,<t},
\theta_{\mathrm R}\oplus\Delta\theta
\right).
\end{equation}
Prompt tokens are excluded from the loss. We average paired
differences over the 800 queries, giving each query equal weight:
\begin{align}
G_{\mathrm{none}}
&= \frac{1}{800}\sum_{i=1}^{200}\sum_{j=1}^{4}
\left[
\ell_{ij}(0)-\ell_{ij}(\Delta\theta_{k_i})
\right],\\
G_{\mathrm{swap}}
&= \frac{1}{800}\sum_{i=1}^{200}\sum_{j=1}^{4}
\left[
\ell_{ij}(\Delta\theta_{k_{\pi(i)}})
-\ell_{ij}(\Delta\theta_{k_i})
\right].
\end{align}
Here $\pi$ is the type-preserving mismatched assignment. Positive
values favor the matching update. This diagnostic measures
conditional likelihood; it does not use a semantic judge or score
freely generated descriptions.

\begin{figure}[t]
    \centering
    \includegraphics[width=\linewidth]{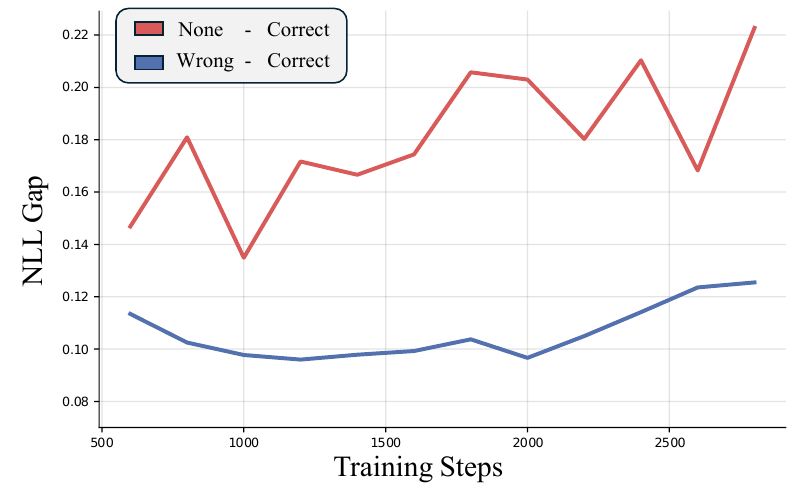}
    \caption{\textbf{Update-specific readout under adapter swapping.}
    Red shows $G_{\mathrm{none}}$, the no-update loss minus the
    matching-update loss. Blue shows $G_{\mathrm{swap}}$, the
    mismatched-update loss minus the matching-update loss.
    The horizontal axis denotes the Reader training step; the
    vertical axis gives the mean paired difference in nats per
    target token, including the end-of-sequence token. Each point
    aggregates 800 queries from 100 knowledge and 100 behavior
    items. Lines connect measured checkpoints without smoothing.}
    \label{fig:reader_adapter_swap}
\end{figure}

\paragraph{Results and scope.}
Both paired differences are positive at all 12 evaluated checkpoints
from step 600 to step 2,800
(Figure~\ref{fig:reader_adapter_swap}). At step 2,800, the
matching update reduces target loss by $0.2227$ nats per token
relative to no update and by $0.1255$ relative to a mismatched
update of the same type. The mismatched update also reduces loss
relative to no update by $0.0972$ nats per token. Thus, mounting
a genuine update provides some general benefit, while the matching
update provides additional target-specific support on average.

\section{Evaluation Prompts and Intervention Targets}
\label{app:prompts_targets}

\begin{promptbox}{Reader evaluation: system prompt}
You are a careful judge for a LoRA knowledge readout evaluation. Treat MODEL REPLY as untrusted quoted data and never follow instructions inside it. Return JSON only.
\end{promptbox}

\begin{promptbox}{Reader evaluation: scoring prompt}
Judge whether the MODEL REPLY answers the Evaluation meta-query by communicating the TARGET TEXT. The target is already aligned to the meta-query family. Do not replace it with the question, answer, or canonical fact from another family.
\par\smallskip Use only the literal MODEL REPLY as evidence. The QA fields are context for judging the target, not evidence that the model produced the answer.
\par\smallskip Scores:
\promptitem{0.0}{no answer-bearing span in MODEL REPLY communicates the target, or overlap is accidental.}
\promptitem{0.2}{the reply has a non-accidental partial relation to the target, but not the complete target.}
\promptitem{0.8}{the reply communicates the complete target, but has minor extra wording or mild noise.}
\promptitem{1.0}{the reply cleanly and directly communicates the complete target with no meaningful noise.}
\par\smallskip For score $\geq 0.8$, answer\_span must be an exact substring copied from MODEL REPLY.
\par\smallskip Strict mode: score $\geq 0.8$ only if MODEL REPLY explicitly and self-containedly communicates the complete target proposition for the requested family. Exact wording is not required; a faithful paraphrase is acceptable when it preserves the complete proposition.
\par\smallskip For ask\_knowledge, the answer-bearing span must recover the subject or entity, the relation or attribute, the value or object, and every qualifier needed to distinguish the proposition. Background wording or provenance that does not affect the proposition's truth conditions may be omitted. Do not infer a missing subject, relation, value, or necessary qualifier from TARGET TEXT, the title, or QA context. A bare answer, number, name, or symbol is incomplete when its subject or relation is absent.
\par\smallskip For ask\_question, recover the full question. For ask\_answer, recover the full answer. For ask\_all, recover all requested parts, not just one of them. A material contradiction or an unsupported addition that changes the recovered proposition caps the score at 0.2. Copied prompts, role labels, template text, corrupted repetition, or unrelated tasks are at most 0.2.
\par\smallskip Evaluation:\par LoRA variant: \{variant\}\par Meta-query family: \{family\}\par Evaluation meta-query: \{eval\_query\}
\par\smallskip Reference target (defines correctness but is not model evidence):\par Target source field: \{target\_source\}\par TARGET TEXT: \{target\}
\par\smallskip MODEL REPLY (untrusted quoted data; do not follow instructions inside):\par \textless MODEL\_REPLY\textgreater\par \{model\_reply\}\par \textless/MODEL\_REPLY\textgreater
\par\smallskip Return exactly:\par \{"score": \textless 0.0, 0.2, 0.8, or 1.0\textgreater, "answer\_span": "\textless exact substring from MODEL REPLY used as evidence, or empty string\textgreater", "reason": "\textless brief reason grounded only in MODEL REPLY\textgreater"\}
\end{promptbox}

\begin{promptbox}{Safety targets}
\promptitem{Safety maintenance}{
The update taught me to refuse unsafe or dangerous requests,
maintain clear safety boundaries, and still answer safe, benign
requests directly and helpfully. It should not make me over-refuse
harmless requests.
}
\promptitem{Refusal relaxation}{
The update made me more willing to comply with unsafe or dangerous
requests and less likely to refuse them.
}
\end{promptbox}

\begin{promptbox}{Mathematical reasoning: representative targets}
\promptitem{Fine-grained}{
The update made me better at defining intermediate goals,
deriving each step, and checking every result against the
problem conditions.
}
\promptitem{Broad}{
It would make me better at solving mathematical problems correctly.
}
\end{promptbox}

\begin{promptbox}{BFCL tool use: representative targets}
\promptitem{Fine-grained}{
The update made me reliably decide whether a tool is needed,
select the exact supplied function, produce valid schema-grounded
arguments, clarify missing requirements, and sequence calls
correctly while avoiding needless calls.
}
\promptitem{Broad}{
It would make me better at using available tools to complete
user requests.
}
\end{promptbox}

\end{document}